# Delay-Optimal Power and Subcarrier Allocation for OFDMA Systems via Stochastic Approximation

Vincent K. N. Lau , *Senior Member, IEEE*, Ying Cui, *Student Member, IEEE*
Department of ECE, The Hong Kong University of Science and Technology

*Abstract*—In this paper, we consider delay-optimal power and subcarrier allocation design for OFDMA systems with $N_F$ subcarriers, $K$ mobiles and one base station. There are $K$ queues at the base station for the downlink traffic to the $K$ mobiles with heterogeneous packet arrivals and delay requirements. We shall model the problem as a $K$-dimensional infinite horizon average reward Markov Decision Problem (MDP) where the control actions are assumed to be a function of the instantaneous Channel State Information (CSI) as well as the joint Queue State Information (QSI). This problem is challenging because it corresponds to a stochastic Network Utility Maximization (NUM) problem where general solution is still unknown. We propose an *online stochastic value iteration* solution using *stochastic approximation*. The proposed power control algorithm, which is a function of both the CSI and the QSI, takes the form of multi-level water-filling. We prove that under two mild conditions in Theorem 1 (One is the stepsize condition. The other is the condition on accessibility of the Markov Chain, which can be easily satisfied in most of the cases we are interested.), the proposed solution converges to the optimal solution almost surely (with probability 1) and the proposed framework offers a possible solution to the general stochastic NUM problem. By exploiting the birth-death structure of the queue dynamics, we obtain a reduced complexity decomposed solution with linear $\mathcal{O}(KN_F)$ complexity and $\mathcal{O}(K)$ memory requirement.

## I. INTRODUCTION

There are plenty of literature on cross-layer optimization of power and subband allocation in OFDMA systems [1], [2]. Yet, all these works focused on optimizing the physical layer performance and the power/subband allocation solutions derived are all functions of the channel state information (CSI) only. On the other hand, real life applications are delay-sensitive and it is critical to consider the delay performance in addition to the conventional physical layer performance in OFDMA cross-layer design. A combined framework taking into account of both queueing delay and physical layer performance is not trivial as it involves both the queueing theory (to model the queue dynamics) and information theory (to model the physical layer dynamics). The first approach converts the delay constraint into average rate constraint using tail probability at large delay regime and solve the optimization problem using information theoretical formulation based on the rate constraint [3]–[5]. While this approach allows potentially simple solution, the derived control policy will be a function of the channel state information (CSI) only. In general, delay-optimal control actions should be a function of both the CSI and queue state information (QSI). In [6], [7], the authors showed that LQHPR policy is delay optimal for multiaccess fading channels. However, the solution utilizes stochastic majorization theory which requires symmetry among the users and is difficult to extend to other situations. In [8]–[10], the authors studied the *queue stability region* of various wireless systems using Lyapunov drift. Under the assumption that all queues are large enough, GPD (Greedy Primal-Dual) algorithm [11] and RT-SPD (Real Time Stochastic Primal-Dual) algorithm [12] are proposed to solve utility-based optimization problem under queueing network stability constraint and average delay constraint separately.

While all the above works addressed different aspects of the delay sensitive resource allocation problem, there are still a number of first order issues to be addressed. In this paper, we consider an OFDMA wireless system with $N_F$ subcarriers, $K$ mobiles and a base station. There are $K$ queues for the mobiles at the base station with heterogeneous arrivals and departures. The delay-optimal power and subcarrier allocation actions, which minimize the average delay of the $K$ MSs under the average total power constraint and subcarrier allocation constraint, are functions of both the CSI and the joint QSI. We shall elaborate the major challenges behind this problem below.

- **The Curse of Dimensionality** A more general approach is to model the problem as a *Markov Decision Problem* (MDP) [13], [14]. However, a primary difficulty in determining the optimal policy using the MDP approach is the huge state space involved. For instance, the state space is exponentially large[1] in the number of users. Hence, brute force solution by value iteration or policy iteration is not applicable due to huge complexity and memory requirement involved.
- **Issues of Convergence in Stochastic NUM Problem** In conventional iterative solutions for deterministic NUM problems, the updates in the iterative algorithms (such as subgradient search) are performed within the coherence time of the CSI (the CSI remains quasi-static during the iteration updates)[2] [15]. In stochastic NUM, however, the updates are performed over the ergodic realizations of the system states, and hence, the convergence proof is challenging (such as GPD in [11] and RT-SPD in [12]). When we consider the delay-optimal problem, the problem is stochastic and the control actions are defined over the ergodic realizations of the system states (CSI,QSI). Therefore, the convergence proof is also quite challenging.

[1]For a system with 4 users, 6 subcarriers, a buffer size of 50 per user and 4 channel states, the system state space contains $4^{4\times 6} \times (50+1)^4$ states, which is already unmanageable.

[2]This poses a serious limitation on the practicality of the distributive iterative solutions because the convergence and the optimality of the iterative solutions are not guaranteed if the CSI changes significantly during the update.

- **Heterogeneous Users** There are some works that obtained a simple delay-optimal policy for multiple access channels by using majorization theory and exploiting symmetry between users [6], [7]. However, such simplifications cannot be extended easily to our case in which users have heterogeneous arrivals and delay requirements.

In this paper, we shall address the above issues by proposing a low-complexity solution to the delay-optimal OFDMA system. To address the open issue concerning the huge complexity involved in solving the $K$-dimensional MDP, we utilize the *stochastic approximation* (SA) techniques [16], [17] to derive a low complexity *online stochastic value iteration* algorithm. We shall show that under some mild conditions, the proposed *online stochastic value iteration* algorithm converges to the optimal solution almost surely (with probability 1). By exploiting the birth-death structure of the queue dynamics, we obtain a reduced complexity decomposed solution with linear $\mathcal{O}(KN_F)$ complexity and $\mathcal{O}(K)$ memory requirement.

## II. SYSTEM MODELS

In this section, we shall elaborate the system model, the OFDMA physical layer model as well as the underlying queueing model. Fig. 1 illustrates the top level system model. There are $K$ user queues at the base station which buffer packets for the $K$ mobile users in the OFDMA system. These $K$ application streams may have different source arrival rates and delay requirements and this corresponds to a heterogeneous user situation. The base station has a *cross-layer scheduler* which takes the CSI and joint QSI as the inputs and produces a power allocation and subcarrier allocation action as outputs.

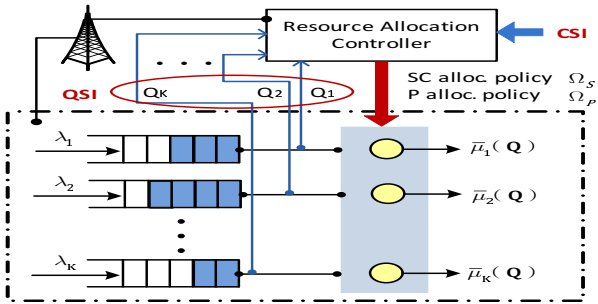

Fig. 1. OFDMA physical layer and queueing model.

### A. OFDMA Physical Layer Model

We consider an OFDMA system with $N_F$ subcarriers over a frequency selective fading channel with $L$ independent multipaths. As a result, the received signal at the $k$-th mobile and $n$-th subcarrier is given by $Y^r_{k,n} = H_{k,n}X^t_{k,n} + Z_{k,n}$, where $X^t_{k,n}$ is the transmitted symbol and $H_{k,n}, Z_{k,n} \sim \mathcal{CN}(0,1)$ are fading coefficient (CSI) and noise respectively. Let $s_{k,n} \in \{0,1\}$ denote the subcarrier allocation index for the $k$-th user at the $n$-th subcarrier. For simplicity, we assume powerful channel coding (such as LDPC) at the transmitter. Furthermore, the mobile receiver has perfect knowledge of CSI. Hence, the maximum achievable data rate of the $k$-th user is given by the mutual information between the channel inputs $\{X_{k,n} : s_{k,n} = 1\}$ and the channel outputs $\{Y_{k,n} : s_{k,n} = 1\}$, which is given by:

$$R_k = \sum_{n=1}^{N_F} s_{k,n} I(X_{k,n}; Y_{k,n}|H_{k,n})$$
$$= \sum_{n=1}^{N_F} s_{k,n} \log\left(1 + p_{k,n}|H_{k,n}|^2\right) \quad (1)$$

### B. Queue Model, System Dynamics and Control Policy

In this paper, the time dimension is partitioned into *scheduling slots* indexed by $t$. Let $\tau$ (second/slot) be the slot duration.

*Assumption 1:* The joint CSI $\mathbf{H}(t) = \{H_{k,n}(t) \forall n, k\}$ remains quasi-static within a scheduling slot and i.i.d. between scheduling slots[3].

There are $K$ queues at the base station transmitter for the $K$ mobiles respectively. Data arrives in packets according to $K$ random arrival processes and each packet is stored in one of the $K$ queues according to its destination. Let $\mathbf{A}(t) = \big(A_1(t), \cdots, A_K(t)\big)$ and $\mathbf{N}(t) = \big(N_1(t), \cdots, N_K(t)\big)$ be the random new packet arrivals at the end of the $t$-th scheduling slot and the packet sizes of the packet in the front of the queues for the $K$ users at the beginning of the $t$-th scheduling slot, respectively.

*Assumption 2:* The arrival process $A_k(t)$ and random packet size $N_k(t)$ are i.i.d. over scheduling slots.

Let $\mathbf{Q}(t) = \big(Q_1(t), \cdots, Q_K(t)\big)$ be the joint QSI of the $K$-user OFDMA system, where $Q_k(t)$ denotes the unfinished work (number of packets) in the $k$-th queue at the beginning of the $t$-th slot. Let $\mathbf{R}(t) = (R_1(t), \cdots, R_K(t))$ be the scheduled data rates (bit/second) of the $K$ users where $R_k(t)$ is given by (1). At the beginning of the $t$-th scheduling slot, the cross-layer scheduler observes the CSI $\mathbf{H}(t)$ and the joint QSI $\mathbf{Q}(t)$ and calculate the scheduled data rate $\mathbf{R}(t)$. We assume the scheduler at the transmitter is causal in the sense that new packet arrivals $\mathbf{A}(t)$ at the $t$-th slot appears after the scheduler's action. Hence, the queue dynamics is given by the following equation.

$$Q_k(t+1) = \min\{\big(Q_k(t) - R_k(t)\tau/N_k(t)\big)^+ + A_k(t), N_Q\} \quad (2)$$

---

[3]The quasi-static assumption is realistic for pedestrian mobility users where the channel coherence time is around 50 ms but typical frame duration is less than 5ms in next generation wireless systems such as WiMAX. On the other hand, we assume the CSI is i.i.d. between slots (as in many other literature) in order to capture first order insights. Similar solution framework can be applied to deal with FSMC fading.

where $x^+ = \max\{x, 0\}$, $N_Q$ denotes the maximum buffer size (number of packets). Thus, the cardinality of the joint QSI is $I = (N_Q + 1)^K$ which grows exponentially with $K$.

For notation convenience, we denote $\boldsymbol{\chi}(t) = (\mathbf{H}(t), \mathbf{Q}(t))$ to be the *global system state* at the $t$-th slot. Given an observed system state realization $\boldsymbol{\chi}$, the transmitter may adjust the transmit power and subcarrier allocation according to a *stationary power control and subcarrier allocation policy* defined below.

*Definition 1:* (*Stationary Power Control and Subcarrier Allocation Policy*) A stationary transmit power and subcarrier allocation policy $\Omega = (\Omega_p, \Omega_s)$ is a mapping from the system state $\boldsymbol{\chi}$ to the power and subcarrier allocation actions. A policy $\Omega$ is called *feasible* if the associated actions satisfy the average total transmit power constraint and the subcarrier assignment constraint. Specifically, $\Omega_p(\boldsymbol{\chi}) = \mathbf{p}$ and $\Omega_s(\boldsymbol{\chi}) = \mathbf{s}$, where $\mathbf{p} = \{p_{k,n}\}$ are the power allocation actions satisfying

$$\sum_{k=1}^{K} \sum_{n=1}^{N_F} \mathbb{E}[p_{k,n}] \leq P_0, \quad p_{k,n} \geq 0 \quad (3)$$

and $\mathbf{s} = \{s_{k,n} \in \{0,1\}\}$ are the subcarrier allocation actions satisfying

$$\sum_{k=1}^{K} s_{k,n} = 1 \quad \forall n \in \{1, N_F\} \quad (4)$$

Note that the $K$ queues are coupled together via the control policy $\Omega$ and the constraints in (3) and (4). The goal of the scheduler at the transmitter is to choose an optimal stationary feasible policy $\Omega^*$ that minimizes the average delays of the $K$ users. Assume that the arrival rate vector falls inside the *stability region* of the system [10]. Given a feasible unichain policy $\Omega$, the induced Markov chain $\{\boldsymbol{\chi}(t)\}$ is ergodic and there exists a unique steady state distribution $\pi_{\boldsymbol{\chi}}$ where $\pi_{\boldsymbol{\chi}}(\boldsymbol{\chi}_0) = \lim_{t \to \infty} \Pr[\boldsymbol{\chi}(t) = \boldsymbol{\chi}_0]$. Hence, by Little's Law [18], the average delay of the $k$-th user under a policy $\Omega$ is given by:

$$\overline{T}_k(\Omega) = \lim_{T \to \infty} \frac{1}{T} \sum_{t=1}^{T} \frac{\mathbb{E}[Q_k(t)]}{\lambda_k} = \frac{\mathbb{E}_{\pi_{\boldsymbol{\chi}}}[Q_k]}{\lambda_k}, \forall k \in \{1, K\} \quad (5)$$

where the $\mathbb{E}_{\pi_{\boldsymbol{\chi}}}$ denotes expectation w.r.t. the underlying measure $\pi$. Similarly, the average transmit power constraint in (3) is given by

$$\overline{P}_{tx}(\Omega) = \lim_{T \to \infty} \frac{1}{T} \sum_{t=1}^{T} \mathbb{E}\left[\sum_{k,n} p_{k,n}(t)\right] = \mathbb{E}_{\pi_{\boldsymbol{\chi}}}\left[\sum_{k,n} p_{k,n}\right] \leq P_0 \quad (6)$$

The average delay is related to the control actions (power and subcarrier allocation) via the packet service rates $\mathbf{R}(t)$ in (1) and the queue dynamics in (2). The delay-optimal scheduler design can be formulated as the following optimization problem[4].

*Problem 1 (Delay-Optimal Policy):* For some constants $\boldsymbol{\beta} = (\beta_1, \cdots, \beta_K)$ ($\beta_k > 0 \, \forall k$), we seek to find a stationary policy $\Omega$ that minimizes

$$J_{\boldsymbol{\beta}}^{\Omega} = \sum_{k=1}^{K} \beta_k \overline{T}_k(\Omega) + \gamma \overline{P}_{tx}(\Omega)$$

$$= \lim_{T \to \infty} \frac{1}{T} \sum_{t=1}^{T} \mathbb{E}\Big[g(\boldsymbol{\chi}(t), \Omega(\boldsymbol{\chi}(t)))\Big] \quad (7)$$

where $g(\boldsymbol{\chi}, \{\mathbf{p}, \mathbf{s}\})$ is the per-stage reward given by:

$$g(\boldsymbol{\chi}, \{\mathbf{p}, \mathbf{s}\}) = \sum_{k} \beta_k \frac{Q_k}{\lambda_k} + \gamma \sum_{k,n} p_{k,n} \quad (8)$$

The positive weighting factors $\boldsymbol{\beta}$ in (7) indicate the relative importance of buffer delay among the $K$ data streams and for each given $\boldsymbol{\beta}$, the solution to (7) corresponds to a point on the Pareto optimal delay tradeoff boundary of a *multi-objective* optimization problem. The constant $\gamma > 0$ is the Lagrange multiplier for the average transmit power constraint in (6).

### III. MARKOV DECISION PROBLEM FORMULATION

In this section, we shall formulate the delay minimization problem in (7) as an infinite horizon Markov Decision Problem (MDP) and discuss the optimality condition.

#### A. MDP Formulation

A stationary control policy $\Omega$ induces a joint distribution for the random process $\{\boldsymbol{\chi}(t)\}$. Since the system queue level $\mathbf{Q}(t)$ evolves according to the system dynamics described in (2) and the arrival, departure and the CSI processes are memoryless, the random process $\boldsymbol{\chi}(t)$ is a Markov chain[5] and hence, the optimization problem in (7) can be modeled as a MDP with transition probability given by[6]:

$$\Pr[\boldsymbol{\chi}(t+1)|\boldsymbol{\chi}(t), \Omega(\boldsymbol{\chi}(t))]$$
$$= \Pr[\mathbf{H}(t+1)|\boldsymbol{\chi}(t), \Omega(\boldsymbol{\chi}(t))] \Pr[\mathbf{Q}(t+1)|\boldsymbol{\chi}(t), \Omega(\boldsymbol{\chi}(t))]$$
$$= \Pr[\mathbf{H}(t+1)] \Pr[\mathbf{Q}(t+1)|\boldsymbol{\chi}(t), \Omega(\boldsymbol{\chi}(t))] \quad (9)$$

As a result, the optimizing policy for the MDP in (7) can be obtained by solving the *Bellman equation* [13] (Page 308) recursively w.r.t. $(\theta, \{V(\boldsymbol{\chi})\})$ as below:

$$\theta + V(\boldsymbol{\chi}) \qquad \qquad \forall \boldsymbol{\chi} \quad (10)$$
$$= \min_{u(\boldsymbol{\chi})} \left[g(\boldsymbol{\chi}, u(\boldsymbol{\chi})) + \sum_{\boldsymbol{\chi}'} \Pr[\boldsymbol{\chi}'|\boldsymbol{\chi}, u(\boldsymbol{\chi})] V(\boldsymbol{\chi}')\right]$$

where $u(\boldsymbol{\chi}) = \Omega(\boldsymbol{\chi}) = \{\mathbf{p}, \mathbf{s}\}$ are the power control and subcarrier allocation actions taken in state $\boldsymbol{\chi}$ and $g(\boldsymbol{\chi}, u(\boldsymbol{\chi}))$ given by (8) is the *per-stage reward* when the current state is $\boldsymbol{\chi}$ and action $u(\boldsymbol{\chi})$ is taken. If there is a $(\theta, \{V(\boldsymbol{\chi})\})$ satisfying (10), then $\theta = J_{\boldsymbol{\beta}}^* = \inf_{\Omega} J_{\boldsymbol{\beta}}^{\Omega}$ is the optimal average reward per stage and the optimizing policy is given by $\Omega^*(\boldsymbol{\chi}) = u^*(\boldsymbol{\chi})$ where $u^*(\boldsymbol{\chi})$ are the optimizing actions of (10) at state $\boldsymbol{\chi}$. Furthermore, since the induced Markov chain $\{\boldsymbol{\chi}(t)\}$ is irreducible for any stationary policy considered, the solution to (10) is unique up to one additive constant.

---

[4]We can apply exactly the same solution framework to the optimization problem which maximizes the physical layer throughput under the average delay constraint and the average power constraint because the Lagrangian function of such problem has the same form as our delay-optimal problem in (7) (both belongs to infinite horizon MDP).

[5]Given current queue state $\mathbf{Q}(t)$, the departure $\mathbf{R}(t)$ (which is determined by the control action $\Omega(\boldsymbol{\chi}(t))$) and arrival $\mathbf{A}(t)$, the future queue level $\mathbf{Q}(t+1)$ is independent of the previous system states.

[6]Although the QSI $\mathbf{Q}(t+1)$ and CSI $\mathbf{H}(t)$ are correlated via the control action $\Omega(\boldsymbol{\chi}(t))$, due to the causality of the control action, $\mathbf{H}(t+1)$ is independent of $\boldsymbol{\chi}(t)$ and $\mathbf{Q}(t+1)$.

### B. Reduced State Bellman Equation

Instead of working on the global state space $\boldsymbol{\chi} = (\mathbf{H}, \mathbf{Q})$, we shall derive a reduced-state Bellman equation from (10) using partitioning of the control policy $\Omega$, which is based on partial system state $\mathbf{Q}$ only. Specifically, we partition a unichain policy $\Omega$ into a collection of actions based on the QSI as follows:

*Definition 2 (Conditional Actions):* Given a control policy $\Omega$, we define $\Omega(\mathbf{Q}) = \{(\mathbf{p}, \mathbf{s}) = \Omega(\boldsymbol{\chi}) : \boldsymbol{\chi} = (\mathbf{Q}, \mathbf{H}) \forall \mathbf{H}\}$ as the collection of actions under a given QSI $\mathbf{Q}$ for all possible CSI $\mathbf{H}$. The policy $\Omega$ is therefore equal to the union of all the conditional actions. i.e. $\Omega = \bigcup_{\mathbf{Q}} \Omega(\mathbf{Q})$.

The following lemma summarizes the main result.

*Lemma 1 (Equivalent Reduced-State Bellman Equation):* The control policy obtained by solving the original Bellman equation in (10) is equivalent to the control policy obtained by solving the following reduced state Bellman equation.

$$\theta + \widetilde{V}(\mathbf{Q}^i) \qquad 1 \leq i \leq I \qquad (11)$$
$$= \min_{\mathbf{u}(\mathbf{Q}^i)} \left[ \widetilde{g}(\mathbf{Q}^i, \mathbf{u}(\mathbf{Q}^i)) + \sum_{\mathbf{Q}^j} \widetilde{f}(\mathbf{Q}^j | \mathbf{Q}^i, \mathbf{u}(\mathbf{Q}^i)) \widetilde{V}(\mathbf{Q}^j) \right]$$

where $\widetilde{V}(\mathbf{Q}) = \mathbb{E}[V(\boldsymbol{\chi})|\mathbf{Q}]$ is the *conditional potential function*, $\mathbf{u}(\mathbf{Q}) = \Omega(\mathbf{Q})$ is the collection of actions under a given QSI $\mathbf{Q}$, $\widetilde{g}(\mathbf{Q}, \mathbf{u}(\mathbf{Q})) = \mathbb{E}[g(\boldsymbol{\chi}, u(\boldsymbol{\chi}))|\mathbf{Q}]$ is the *conditional per-stage reward*, $\widetilde{f}(\mathbf{Q}^j | \mathbf{Q}^i, \mathbf{u}(\mathbf{Q}^i)) = \mathbb{E}\left[\Pr[\mathbf{Q}^j | \boldsymbol{\chi}^i, u(\boldsymbol{\chi}^i)] | \mathbf{Q}^i\right]$ is the *conditional average transition kernel*.

*Proof:* Please refer to Appendix A for the proof. ∎

A solution to (11) is still very complex due to the huge dimensionality of the state space ($I$ is exponential in $K$) and brute force *value iteration* or *policy iteration* [19] has exponential memory size requirement. As a result, it is desirable to obtain an online and low-complexity solution for the problem.

## IV. GENERAL SOLUTION TO THE DELAY OPTIMAL PROBLEM

In this section, we shall derive a low complexity (but optimal) solution by proposing an online value iteration to solve the reduced state Bellman equation in (11). We shall also establish technical conditions for almost-sure convergence of the online value iteration.

### A. Online value iteration via Stochastic Approximation

We shall propose an online *sample-path-based* iterative learning algorithm to estimate the performance potential and the control policy. Define a vector mapping $\mathbf{T} : R^I \to R^I$ with the $i$-th component mapping $(1 \leq i \leq I)$ as

$$T_i(\widetilde{\mathbf{V}}) = \min_{\mathbf{u}(\mathbf{Q}^i)} \left\{ \widetilde{g}(\mathbf{Q}^i, \mathbf{u}(\mathbf{Q}^i)) + \sum_{\mathbf{Q}^j} \widetilde{f}(\mathbf{Q}^j | \mathbf{Q}^i, \mathbf{u}(\mathbf{Q}^i)) \widetilde{V}(\mathbf{Q}^j) \right\}$$
(12)

Since the potential is unique up to a constant, we could set $T_i(\widetilde{\mathbf{V}}) - \widetilde{V}(\mathbf{Q}^i) = J_\beta$ for some reference state $\mathbf{Q}^i$ ($1 \leq i \leq I$). Let $t = \{0, 1, 2, ...\}$ be the slot index and $\{\mathbf{Q}(0), \cdots, \mathbf{Q}(t), \cdots\}$ be the *sample-path*, i.e. the corresponding realizations of the system states. To perform the online value iteration, we divide the sample-path into regenerative periods. A regenerative period is defined as the minimum interval that each $\mathbf{Q}$ state is visited at least once. Let $l_k(i)$ and $\widehat{\mathbf{V}}_k$ be the times that $\mathbf{Q}^i$ is visited and the estimated performance potential in the $k$-th regenerative period respectively. Let $n_0 = 0$, $n_{k+1} = \min\{t + 1 : t > n_k, \min_i l_k(i) = 1\}$ for $k \geq 0$. Then the sample path in the $k$-th regenerative period is $\{\mathbf{Q}(n_k), \cdots, \mathbf{Q}(n_{k+1} - 1)\}$. At the beginning of the $k$-th regenerative period ($n_k \leq t \leq n_{k+1} - 1$), initialize the following dummy variables as $S_{\widehat{g}_k}(i) = 0$, $S_{\widehat{V}_k}(i) = 0$ and $l_k(i) = 0$. Within the $k$-th regenerative period, we adopt policy $\Omega_k$. After observing the queue state $Q(t+1)$ at the end of the $t$-th slot, update the following metric of the visited queue state $Q(t)$ according to

$$\begin{cases} S_{\widehat{g}_k}(i) &= S_{\widehat{g}_k}(i) + g\big(\boldsymbol{\chi}(t), \Omega_k(\boldsymbol{\chi}(t))\big) \\ S_{\widehat{V}_k}(i) &= S_{\widehat{V}_k}(i) + \widehat{V}_k\big(\mathbf{Q}(t+1)\big) \quad \text{if } \mathbf{Q}^i = \mathbf{Q}(t) \\ l_k(i) &= l_k(i) + 1 \end{cases} \quad (13)$$

At the end of the $k$-th regenerative period, using stochastic approximation algorithm [17], we update the estimated potential for the $(k+1)$-th regenerative period, which is

$$\widehat{V}_{k+1}(\mathbf{Q}^i) = \widehat{V}_k(\mathbf{Q}^i) + \epsilon_k Y_k(\mathbf{Q}^i), \ 1 \leq i \leq I \qquad (14)$$

where

$$Y_k(\mathbf{Q}^i) = \frac{S_{\widehat{g}_k}(i)}{l_k(i)} - \Big(\frac{S_{\widehat{g}_k}(I)}{l_k(I)} + \frac{S_{\widehat{V}_k}(I)}{l_k(I)} - \widehat{V}_k(\mathbf{Q}^I)\Big) + \frac{S_{\widehat{V}_k}(i)}{l_k(i)} - \widehat{V}_k(\mathbf{Q}^i)$$

and $\epsilon_k$ is the step size of the stochastic approximation algorithm and $\mathbf{Q}^I$ is the reference state[7]. Accordingly, we update the policy for the $(k+1)$-th regenerative period, which is given by

$$\Omega_{k+1} = \arg\min \mathbf{T}(\widehat{\mathbf{V}}_{k+1}) \qquad (15)$$

Therefore, we could construct an online value iteration algorithm as below.

---

**Algorithm 1: Online Value Iteration**

· Step 1 (**Initialization**): Set $t = 0$, and start the system at an initial state $\mathbf{Q}(0)$. Set $k = 0$, initialize the potential $\widehat{\mathbf{V}}_0$ and policy $\Omega_0 = \arg\min \mathbf{T}(\widehat{\mathbf{V}}_0)$ in the 0-th regenerative period.

· Step 2 (**Online Potential Estimation**): At the beginning of the $k$-th regenerative period, set $S_{\widehat{g}_k}(i) = 0$, $S_{\widehat{V}_k}(i) = 0$ and $l_k(i) = 0 \, \forall i$. Run the system with policy $\Omega_k$ to $n_{k+1} - 1$ and accumulate the information of the visited $\mathbf{Q}$ from slot to slot according to (13). At $n_{k+1} - 1$, update the estimated potential $\widehat{V}_{k+1}$ for the $(k+1)$-th regenerative period according to (14).

· Step 3 (**Online Policy Improvement**): Update the policy $\Omega_{k+1}$ for the $(k+1)$-th regenerative period according to (15).

· Step 4 (**Termination**): If $\|\widehat{V}_{k+1} - \widehat{V}_k\| < \delta_v$, stop; otherwise, set $k := k + 1$ and go to Step 2.

---

Fig. 2 illustrates the online value iteration algorithm with an example.

---

[7]Without loss of generality, we set the state that all $K$ buffers are empty as $\mathbf{Q}^I$ and initialize $\widehat{V}_0(\mathbf{Q}^I) = 0$.

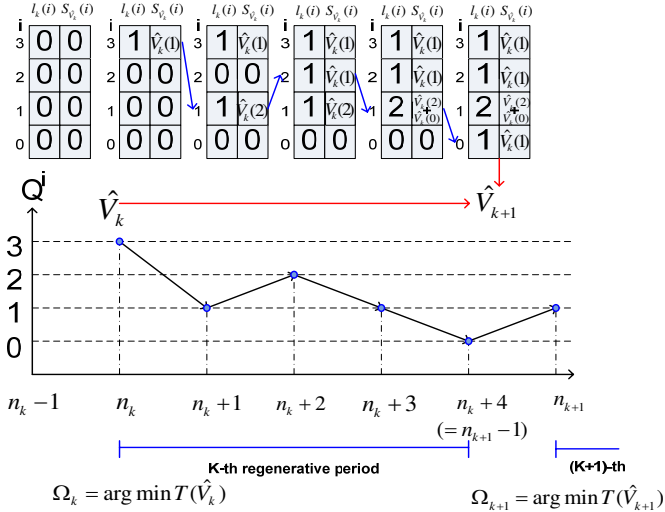

Fig. 2. Illustration of online sample path based potential learning algorithm. $K = 2$, $N_Q = 1$, $I = (N_Q+1)^2 = 4$, the four joint queue states are 00, 01, 10, 11, which are denoted as 0, 1, 2, 3 for simplicity. The sample path in the $k$-th regenerative period ($n_k \leq t \leq n_{k+1} - 1$) is $\{3, 1, 2, 1, 0, 1\}$. At the beginning of the $n_k$-th slot, set $S_{\widehat{g}_k}(i) = 0$, $S_{\widehat{V}_k}(i) = 0$ and $l_k(i) = 0$. In the $k$-th regenerative period, adopt policy $\Omega_k$ and update $S_{\widehat{g}_k}(i)$, $S_{\widehat{V}_k}(i)$ and $l_k(i)$ according to (13) from slot to slot. At the end of the $(n_{k+1}-1)$-th slot, update the potential $\widehat{V}_{k+1}(\mathbf{Q}^i)$ and policy $\Omega_{k+1}$ for the $(k+1)$-th regenerative period according to (14) and (15).

*Remark 1 (Comparison to the deterministic NUM):* In conventional iterative solutions for deterministic NUM [15], the iterative updates (with message exchange)[8] are performed within the CSI coherence time and hence, this limits the number of iterations and the performance. However, in the proposed online algorithm, the updates in the iteration steps evolve in the same time scale as the CSI and QSI. Hence, the algorithm could converge to a better solution because the number of iterations is no longer limited by the coherence time of CSI.

### B. Convergence Analysis

In this part, we shall establish the technical proof about the almost sure convergence of the online value iteration in (14). Assume the sequence of step size $\{\epsilon_k\}$ is chosen such that it satisfies the following stepsize conditions:

$$\sum_k \epsilon_k = \infty, \quad \epsilon_k \geq 0, \quad \epsilon_k \to 0, \quad \sum_k \epsilon_k^2 < \infty \quad (16)$$

Let $\mathbb{E}_k$ and $\Pr_k$ denote the expectation and probability conditioned on the $\sigma$-algebra $\mathcal{F}_k$, generated by $\{\widehat{\mathbf{V}}_0, \mathbf{Y}_i, i < k\}$. Define $\delta M_k(\mathbf{Q}^i) \triangleq Y_k(\mathbf{Q}^i) - \mathbb{E}_k[Y_k(\mathbf{Q}^i)]$ and

$$\mathbf{A}_{k-1} \triangleq P_{\Omega_k} \epsilon_{k-1} + (1 - \epsilon_{k-1})\mathbf{I}$$
$$\mathbf{B}_{k-1} \triangleq P_{\Omega_{k-1}} \epsilon_{k-1} + (1 - \epsilon_{k-1})\mathbf{I}$$

where $P_{\Omega_k}$ and $\mathbf{I}$ are $I \times I$ transition matrix under policy $\Omega_k$ and identity matrix. The convergence of the online value iteration algorithm is summarized in the following theorem.

[8]Since the iterations within a CSI coherence time involve explicit message passing, there is processing and signaling overhead per iteration and this limits the total number of iterations within a CSI coherence time.

*Theorem 1 (Convergence of Online Value Iteration):* Assume $\delta \mathbf{M}_k$ is bounded $\forall k$ and the sequence of step size $\{\epsilon_k\}$ satisfies the conditions in (16). Assume that for every set of control policies $\Omega_0, \cdots, \Omega_m$, there exist a $\delta_m = \mathcal{O}(\epsilon_m) > 0$ and some positive integer $m$ such that

$$[\mathbf{A}_m \cdots \mathbf{A}_1]_{iI} \geq \delta_m, \quad [\mathbf{B}_m \cdots \mathbf{B}_1]_{iI} \geq \delta_m, \quad 1 \leq i \leq I \quad (17)$$

where $[\cdot]_{iI}$ denotes the element of the $i$-th row and $I$-th column of the corresponding $I \times I$ matrix and $\mathbf{e} = [1, \cdots, 1]^T$ is the $I \times 1$ unit vector. For an arbitrary initial potential vector $\widehat{\mathbf{V}}_0$, we have $\lim_{k \to \infty} \widehat{\mathbf{V}}_k = \widehat{\mathbf{V}}_\infty$ w.p.1, where the steady state potential $\widehat{\mathbf{V}}_\infty$ satisfies:

$$(T_I(\widehat{\mathbf{V}}_\infty) - \widehat{V}_\infty(\mathbf{Q}^I))\mathbf{e} + \widehat{\mathbf{V}}_\infty = \mathbf{T}(\widehat{\mathbf{V}}_\infty) \quad (18)$$

Furthermore, the optimal reward of the delay optimal problem is $J_\beta^* = T_I(\widehat{\mathbf{V}}_\infty) - \widehat{V}_\infty(\mathbf{Q}^I)$ and the optimal stationary policy is $\Omega^* = \arg\min \mathbf{T}(\widehat{\mathbf{V}}_\infty)$.

*Proof:* Please refer to Appendix B. ■

*Remark 2:* Note that $\mathbf{A}_k$ and $\mathbf{B}_k$ are related to an equivalent transition matrix of the underlying Markov chain. (17) simply means that state $\mathbf{Q}^I$ is accessible from all the $\mathbf{Q}$ states after some finite number of transition steps. This is a very mild condition and will be satisfied in most of the cases we are interested.

*Remark 3:* Note that (18) is equivalent to the reduced state Bellman equation in (11). As a result, the converged solution of (18) corresponds to the solution of (11).

## V. APPLICATION TO OFDMA SYSTEMS WITH POISSON ARRIVAL

In this section, we shall illustrate the usage of the online value iteration to minimize the average weighted delay of OFDMA systems under Poisson packet arrival and exponential distributed packet size.

*Assumption 3:* The arrival process $A_k(t)$ is i.i.d. over scheduling slots following Poisson distribution with average arrival rate $\mathbb{E}[A_k] = \lambda_k$. The random packet size $N_k(t)$ is i.i.d. over scheduling slots following an exponential distribution with mean packet size $\overline{N}_k$.

*Assumption 4:* The slot duration $\tau$ is sufficiently small compared with the average packet interarrival time as well as conditional average packet service time[9], i.e. $\lambda_k \tau \ll 1$ and $\mu_k(\chi)\tau \ll 1$.

Conditioned on the system state $\chi$, the conditional mean departure rate of user $k$ is given by $\mu_k(\chi) = \mathbb{E}[R_k(\chi)/N_k|\chi] = R_k(\chi)/\overline{N}_k$. Thus, the conditional probability (conditioned on the current system state realization $\chi(t)$) of a packet departure event at the $t$-th slot is given by

$$\Pr\left[\frac{N_k(t)}{R_k(t)} < \tau | \chi(t), \Omega(\chi(t))\right] = \Pr\left[\frac{N_k(t)}{\overline{N}_k} < \mu_k(\chi(t))\tau\right]$$
$$= 1 - \exp(-\mu_k(\chi(t))\tau) \approx \mu_k(\chi(t))\tau$$

where the "≈" is due to Assumption 4. Note that under Assumption 4, the probability for simultaneous arrival, departure of two or more packets from the same queue or different queues and simultaneous arrival as well as departure in a slot are $\mathcal{O}((\lambda_k\tau)^2)$, $\mathcal{O}((\mu_k(\chi(t))\tau)^2)$ and $\mathcal{O}(\lambda_k\tau\mu_k(\chi(t))\tau)$ respectively, which are asymptotically negligible. Therefore, the vector queue dynamics are given by

$$\widetilde{f}[(Q_1^i,\cdots,(Q_k^i+1)_{\wedge N_Q},\cdots,Q_K^i)|\mathbf{Q}^i,\mathbf{u}(\mathbf{Q}^i)] \approx \lambda_k\tau,\ \forall k$$
$$\widetilde{f}[(Q_1^i,\cdots,(Q_k^i-1)^+,\cdots,Q_K^i)|\mathbf{Q}^i,\mathbf{u}(\mathbf{Q}^i)] \approx \overline{\mu}_k(\mathbf{Q}^i)\tau,\ \forall k$$
$$\widetilde{f}[\mathbf{Q}^i|\mathbf{Q}^i,\mathbf{u}(\mathbf{Q}^i)] \approx 1 - \sum_k \left(\lambda_k\tau + \overline{\mu}_k(\mathbf{Q}^i)\tau\right) \quad (19)$$

where $x_{\wedge N_Q} = \min\{x,N_Q\}$, $\mathbf{Q}^i = (Q_1^i,\cdots,Q_K^i)$, and $\overline{\mu}_k(\mathbf{Q}) = \mathbb{E}[\mu_k(\chi)|\mathbf{Q}]$. In what follows, we shall discuss the optimal solution and asymptotically optimal solution with only linear complexity and memory requirement of the OFDMA system with the conditional transition kernel given by (19).

[9]This is a mild assumption which could be justified in many applications. For example, in WiMAX, a frame duration is around 2ms while the target queueing delay for applications (such as video streaming) is around 200ms or more.

### A. Optimal Solution

Given (19), the optimization objective in the R.H.S. of Bellman equation in (11) becomes

$$\mathbb{E}\Big[\gamma\sum_{k,n} p_{k,n}(\chi^i) - \sum_{k=1}^K \Delta_k\widetilde{V}(\mathbf{Q}^i)\frac{\tau}{\overline{N}_k} \quad (20)$$
$$\cdot\Big(\sum_{n=1}^{N_F} s_{k,n}\log(1+p_{k,n}|H_{k,n}|^2)\Big)|\mathbf{Q}^i\Big]$$

where $\Delta_k\widetilde{V}(\mathbf{Q}^i) \triangleq \widetilde{V}(Q_1^i,\cdots,Q_K^i) - \widetilde{V}(Q_1^i,\cdots,[Q_k^i-1]^+,\cdots,Q_K^i)$. Using standard optimization techniques, the closed-form solution during the policy improvement step (15) in the online-value iteration algorithm is summarized below.

*Lemma 2:* (*Closed-Form Power Control and Subcarrier Allocation of Online Policy Improvement*) Under the above setup, the optimizing power control and subcarrier allocation actions in the policy improvement step (15) for given CSI $\mathbf{H}$ and QSI $\mathbf{Q}$ are given by:

$$p_{k,n}(\mathbf{H},\mathbf{Q}) = s_{k,n}(\mathbf{H},\mathbf{Q})\Big(\frac{\frac{\tau}{\overline{N}_k}\Delta_k\widetilde{V}(\mathbf{Q})}{\gamma} - \frac{1}{|H_{k,n}|^2}\Big)^+ \quad (21)$$
$$s_{k,n}(\mathbf{H},\mathbf{Q}) = \begin{cases} 1, & \text{if } X_{k,n} = \max_j\{X_{j,n}\} > 0 \\ 0, & \text{otherwise} \end{cases} \quad (22)$$

where $\gamma$ satisfies (6) and $X_{k,n} = \frac{\tau}{\overline{N}_k}\Delta_k\widetilde{V}(\mathbf{Q})\log\Big(1+|H_{k,n}|^2\big(\frac{\frac{\tau}{\overline{N}_k}\Delta_k\widetilde{V}(\mathbf{Q})}{\gamma} - \frac{1}{|H_{k,n}|^2}\big)^+\Big) - \gamma\big(\frac{\frac{\tau}{\overline{N}_k}\Delta_k\widetilde{V}(\mathbf{Q})}{\gamma} - \frac{1}{|H_{k,n}|^2}\big)^+$.

*Proof:* Due to the combinatorial nature of the subcarrier allocation in the OFDMA system, i.e. $s_{k,n} \in \{0,1\}$ and (4), finding the optimal solution by brute force exhaustive search requires exponential complexity [20]. By applying continuous relaxation technique [20], [21], the original integer programming problem can be relaxed to a convex optimization problem and then the standard convex optimization techniques can be applied. It can be shown that the original problem and the relaxed problem share the same optimal solution in general. We omit the detailed proof here due to page limit. ∎

*Remark 4:* (*Structure of the Optimal Power Control and Subcarrier Allocation*) The optimal power control and the subcarrier allocation solution in Lemma 2 are both functions of CSI and QSI where they depend on the QSI indirectly via the potential function $\{\widetilde{V}(\mathbf{Q})\}$. For the power control solution, it has the form of *multi-level water-filling* where the power is allocated according to the CSI across subcarriers but the water-level is adaptive to the QSI. Similarly, the subcarrier is selected according to the metric $X_{k,n}$ which depends on both the CSI and the QSI.

While the online value iteration is much simplified compared with the brute-force MDP solutions, it still suffers from exponentially large memory requirement and computational complexity for storing and computing the potential vector. In the next subsection, we shall exploit the birth-death dynamics and derive an asymptotically optimal solution with linear $\mathcal{O}(KN_F)$ computational complexity and $\mathcal{O}(K)$ memory requirement.

### B. Asymptotically Optimal Solution

We first define a simplified subcarrier allocation policy below and summarize an important structural result of the Bellman equation in (11) under the simplified policy.

*Definition 3:* [CSI-Only Subcarrier Allocation Policy] A *CSI-only subcarrier allocation policy* is defined as $\widetilde{\Omega}_s(\mathbf{H}) = \{\tilde{s}_{k,n}(\mathbf{H}) \in \{0,1\}|\sum_{k=1}^K \tilde{s}_{k,n} = 1\ \forall n\}$.

*Theorem 2 (Additive Property of the Potential Function):* Under the average power constraint in (3) and the *CSI-only subcarrier allocation policy*, the solution of the Bellman equation in (11) can be expressed into the form $\widetilde{V}(\mathbf{Q}) = \sum_k \widetilde{V}_k(Q_k)$ and $\theta = \sum_k \theta_k$, where $\{\widetilde{V}_k(Q_k),\theta_k\}$ is the solution of the $k$-th user's reduced Bellman equation:

$$\theta_k = \min_{\mathbf{u}_k(Q_k)}\Big\{\widetilde{g}_k(Q_k,\mathbf{u}_k(Q_k)) + \lambda_k\tau\Delta\widetilde{V}_k(Q_k+1)$$
$$- \overline{\mu}_k(Q_k)\tau\Delta\widetilde{V}_k(Q_k)\Big\} \quad (23)$$

where $\mathbf{u}_k(Q_k) = \{(\tilde{p}(\mathbf{H},Q_k),\tilde{s}(\mathbf{H})) : (\mathbf{H},Q_k)\forall\mathbf{H}\}$ is the set of the conditional control actions, $\overline{\mu}_k(Q_k) = \mathbb{E}\big[\sum_{n=1}^{N_F} \tilde{s}_{k,n}(\mathbf{H})\log(1+\tilde{p}_{k,n}(\mathbf{H},Q_k)|H_{k,n}|^2)|Q_k\big]/\overline{N}_k$ is the conditional average service rate and $\Delta\widetilde{V}_k(Q_k) = \widetilde{V}_k(Q_k) - \widetilde{V}_k([Q_k-1]^+)$ is the potential increment of the $k$-th queue.

*Proof:* Please refer to Appendix C for the proof. ∎

As a result of Theorem 2, if we restrict the subcarrier allocation policy to the *CSI-only subcarrier allocation policy*, then the potential function $\{\widetilde{V}(\mathbf{Q})\}$ of the joint QSI for the original MDP problem can be decomposed into $K$ individual potential functions $\{\widetilde{V}_k(Q_k)\}$ of the individual QSI for the $K$ individual MDP's. This could substantially simplify the potential estimation, the convergence speed as well as the memory requirement. In particular, to satisfy the condition of Theorem 2, we modify the optimal subcarrier allocation solution in Lemma 2 to a *CSI-only policy* which is given by

$$\tilde{s}_{k,n}(\mathbf{H}) = \begin{cases} 1, & \text{if } |H_{k,n}|^2 = \max_j\{|H_{j,n}|^2\} \\ 0, & \text{otherwise} \end{cases} \quad (24)$$

While using (24) will result in some loss of optimality in strict sense, we shall show in the following Corollary that the solution is indeed asymptotically optimal.

*Corollary 1 (Reduced Complexity Online Value Iteration):*
The following reduced complexity online value iteration algorithm has $\mathcal{O}(KN_F)$ complexity and $\mathcal{O}(K)$ memory requirement. It converges almost surely to a solution which is asymptotically optimal for sufficiently large $K$[10].

---

Algorithm 2: **Reduced Complexity Online value iteration**

· Step 1 (**Initialization**): For each user $k = 1 : K$, initialize the potential $\widehat{\mathbf{V}}_k^0$ and policy $\Omega_k^0 = \arg\min \mathbf{T}(\widehat{\mathbf{V}}_k^0)$ in the 0-th regenerative period. Set $t = 0$, and start the OFDMA at an initial QSI state $Q_k(0)$ for each user.

· Step 2 (**Online Potential Estimation**): For each user $k = 1 : K$ in its $l_k$-th regenerative period, set $S_{\widehat{g}_k^{l_k}}(Q_k) = 0$, $S_{\widehat{V}_k^{l_k}}(Q_k) = 0$ and $j_k^{l_k}(Q_k) = 0\ \forall Q_k$. Perform power control and subcarrier allocation according to the policy $\Omega_k^{l_k}$, and accumulate the information of the visited $Q_k$ for each user $k$ if $Q_k = Q_k(t)$ according to

$$\begin{cases} S_{\widehat{g}_k^{l_k}}(Q_k) &= S_{\widehat{g}_k^{l_k}}(Q_k) + g_k\big(\chi_k(t), \Omega_k^{l_k}(\chi_k(t))\big) \\ S_{\widehat{V}_k^{l_k}}(Q_k) &= S_{\widehat{V}_k^{l_k}}(Q_k) + \widehat{V}_k^{l_k}(Q_k(t+1)) \\ j_k^{l_k}(Q_k) &= j_k^{l_k}(Q_k) + 1 \end{cases}$$

If the current slot corresponds to the end of any of the $K$ regenerative period, update the corresponding estimated potential $\widehat{V}_k^{l_k}$ for its next regenerative period according to ($0 \leq Q_k \leq N_Q$)

$$\widehat{V}_k^{l_k+1}(Q_k) = \widehat{V}_k^{l_k}(Q_k) + \epsilon_{l_k}\bigg[ \frac{S_{\widehat{g}_k^{l_k}}(Q_k)}{j_k^{l_k}(Q_k)} + \frac{S_{\widehat{V}_k^{l_k}}(Q_k)}{j_k^{l_k}(Q_k)} $$
$$- \Big( \frac{S_{\widehat{g}_k^{l_k}}(Q_k^I)}{l_k(Q_k^I)} + \frac{S_{\widehat{V}_k^{l_k}}(Q_k^I)}{l_k(Q_k^I)} - \widehat{V}_k^{l_k}(Q_k^I) \Big) - \widehat{V}_k^{l_k}(Q_k) \bigg]$$

· Step 3 (**Online Policy Improvement**): For each user $k = 1 : K$, if the current slot corresponds to the end of any of the $K$ regenerative periods, then update the policy for its next regenerative period at the BS according to (24) and

$$\tilde{p}_{k,n}(\mathbf{H}, Q_k) = \tilde{s}_{k,n}(\mathbf{H})\Big( \frac{\frac{\tau}{\overline{N}_k}\Delta \widehat{V}_k^{l_k+1}(Q_k)}{\gamma} - \frac{1}{|H_{k,n}|^2} \Big)^+ \quad (25)$$

· Step 4 (**Termination**): If $||\widehat{V}_k^{l_k+1} - \widehat{V}_k^{l_k}|| < \delta_v\ \forall k$, stop; otherwise, set $l_k := l_k + 1$ and go to Step 2.

---

*Proof:* Please refer to Appendix D for the proof. ∎

## VI. SIMULATION RESULTS AND DISCUSSIONS

In this section, we shall compare our proposed optimal and reduced complexity solutions by online value iteration via stochastic approximation to the delay optimal problem for the system with Poisson arrival and exponential packet size with three reference baselines. Baseline 1 refers to a throughput optimal policy[11], namely the *Modified Largest Weighted Delay First (M-LWDF)* [22]. Baseline 2 refers to the *Real Time Stochastic Primal Dual (RT-SPD)* algorithm [12]. Baseline 3 refers to the *Round Robin Scheduling*, in which different users are served in TDMA fashion with equally allocated time slots and water-filling power allocation across the subcarriers. In the simulation, we assume there are 64 subcarriers with total bandwidth 10MHz, and the number of independent subbands $N_F$ is 4. The scheduling slot duration $\tau$ is 5ms. The buffer size $N_Q$ is 10. The average packet arrival rate of Poisson process $\lambda_k$ is 20 packet/s.

---

[10]As we scale up $K$, we assume the transmit total power $P_0$ is sufficiently large so that $(\lambda_1, \cdots, \lambda_K)$ still remains in the stability region of the system

[11]Throughput optimal policy means that it shall stabilize the queue whenever the arrival rate vector falls within the stability region.

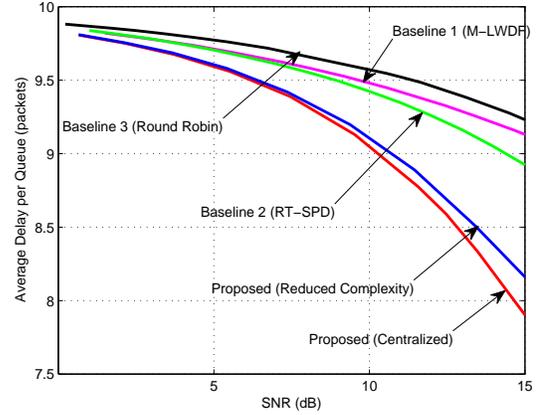

Fig. 3. Average delay per queue versus SNR. The number of users $K = 2$, the buffer size $N_Q = 10$, the mean packet size $\overline{N}_k = 305.2$ Kbyte/pck, the average arrival rate $\lambda_k = 20$ pck/s, the queue weight $\beta_1 = \beta_2 = 1$. The packet drop rate of the proposed schemes are 5%, while the packet drop rate of the Baseline 1 (M-LWDF), Baseline 2 (RT-SPD) and Baseline 3 (Round Robin) are 5%, 5%, 6% respectively.

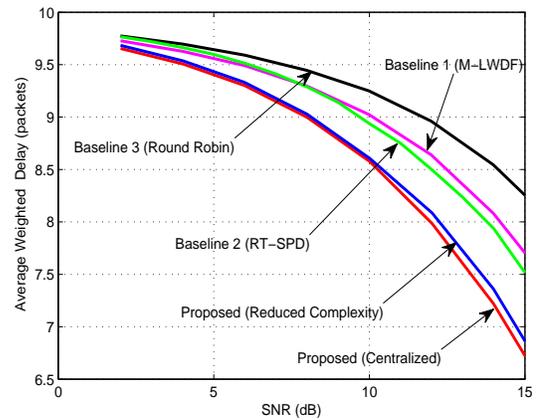

Fig. 4. Average weighted delay versus SNR. The number of users $K = 2$, the buffer size $N_Q = 10$, the mean packet size $\overline{N}_k = 305.2$ Kbyte/pck, the average arrival rate $\lambda_k = 20$ pck/s, the queue weight $\beta_1 = 1$, $\beta_2 = 4$. The packet drop rate of the proposed schemes are 4%, while the packet drop rate of the Baseline 1 (M-LWDF), Baseline 2 (RT-SPD) and Baseline 3 (Round Robin) are 4%, 6%, 6% respectively.

Fig. 3 illustrates the average delay per queue versus SNR of 2 users with equal queue weight. It can be observed that both the optimal solution and reduced complexity solution have significant gain compared with three baselines (e.g. more than 5 dB gain when average delay per queue is less than 9 packets). In addition, the delay performance of the reduced complexity solution, which is asymptotically optimal in large number of users, is very close to the performance of the optimal solution even in 2 user-case. Fig. 4 depicts the average weighted delay versus SNR of 2 heterogeneous users with different queue weight. The average weighted delay of the reduced complexity solution is close to that of the optimal solution as well. Therefore, the proposed reduced complexity solution with linear $\mathcal{O}(KN_F)$ memory requirement and computational complexity as well as near optimal performance is of great practical significance.

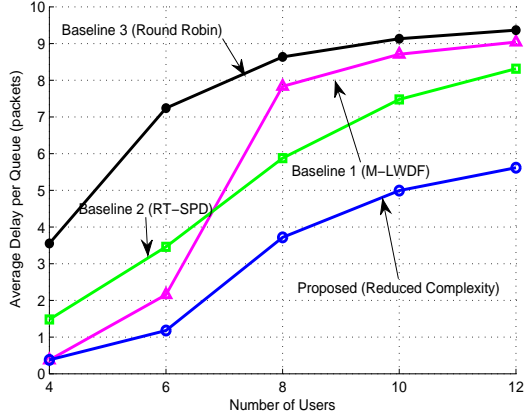

Fig. 5. Average delay per queue versus the number of users. The buffer size $N_Q = 10$, the mean packet size $\overline{N}_k = 39$ Kbyte/pck, the average arrival rate $\lambda_k = 20$ pck/s, the queue weight $\beta_k = 1$ at a transmit SNR= 10dB. The packet drop rate of the proposed scheme is 4%, while the packet drop rate of the Baseline 1 (M-LWDF), Baseline 2 (RT-SPD) and Baseline 3 (Round Robin) are 4%, 4%, 6% respectively.

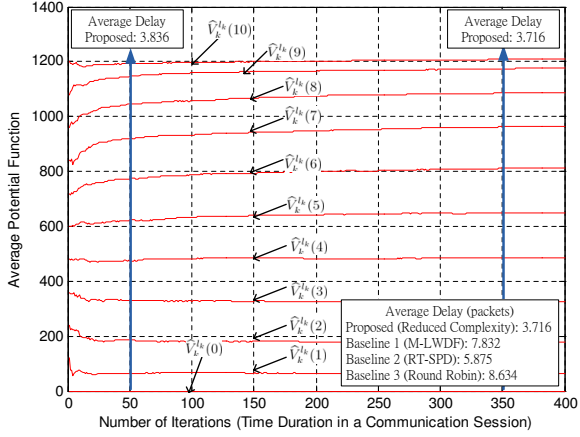

Fig. 6. Illustration of convergence property. Potential function versus the number of iterations. The number of users $K = 8$, the buffer size $N_Q = 10$, the mean packet size $\overline{N}_k = 39$ Kbyte/pck, the average arrival rate $\lambda_k = 20$ pck/s, the queue weight $\beta_k = 1$ at a transmit SNR= 10dB. The packet drop rate of the proposed scheme is 1%, while the packet drop rate of the Baseline 1 (M-LWDF), Baseline 2 (RT-SPD) and Baseline 3 (Round Robin) are 1%, 1%, 4% respectively.

Fig. 5 illustrates the average delay per queue of the reduced complexity solution versus the number of users with equal queue weight at a transmit SNR= 10dB. It is obvious that the reduced complexity solution has great gain in delay over the three baselines in the whole user region.

Fig. 6 illustrates the convergence property of the proposed reduced complexity algorithm. We plot the average potential function of 8 users versus the number of iterations at a transmit SNR= 10dB. It can be seen that the reduced complexity algorithm converges quite fast. The average delay corresponding to the potential function at the 50-th iteration is 3.8 packets, which is much smaller than the other baselines.

## VII. SUMMARY

In this paper, we propose a low-complexity solution to the delay-optimal power and subcarrier allocation design for OFDMA systems. We model the problem as a $K$-dimensional infinite horizon average reward MDP with the control actions based on CSI and joint QSI. We derive the equivalent reduced state Bellman equation and propose an *online stochastic value iteration* solution using *stochastic approximation*. We prove that under some mild conditions[12], the proposed solution converges to the optimal solution almost surely (with probability 1). By exploiting the birth-death structure of the queue dynamics in the Poisson arrivals, we obtain a reduced complexity decomposed solution with linear $\mathcal{O}(KN_F)$ complexity and $\mathcal{O}(K)$ memory requirement.

## APPENDIX

### APPENDIX A: PROOF OF LEMMA 1

$$\theta + V(\mathbf{H}, \mathbf{Q}^i) \qquad \forall \mathbf{H}, 1 \leq i \leq I$$
$$= \min_{u(\mathbf{H}, \mathbf{Q}^i)} \left[ g((\mathbf{H}, \mathbf{Q}^i), u(\mathbf{H}, \mathbf{Q}^i)) \right.$$
$$\left. + \sum_{(\mathbf{H}', \mathbf{Q}^j)} \Pr[(\mathbf{H}', \mathbf{Q}^j) | (\mathbf{H}, \mathbf{Q}^i), u(\mathbf{H}, \mathbf{Q}^i)] V(\mathbf{H}', \mathbf{Q}^j) \right]$$
$$\stackrel{(a)}{=} \min_{u(\mathbf{H}, \mathbf{Q}^i)} \left[ g((\mathbf{H}, \mathbf{Q}^i), u(\mathbf{H}, \mathbf{Q}^i)) \right.$$
$$\left. + \sum_{\mathbf{Q}^j} \Pr[\mathbf{Q}^j | (\mathbf{H}, \mathbf{Q}^i), u(\mathbf{H}, \mathbf{Q}^i)] \left( \sum_{\mathbf{H}'} \Pr[\mathbf{H}'] V(\mathbf{H}', \mathbf{Q}^j) \right) \right]$$
$$\stackrel{(b)}{=} \min_{u(\mathbf{H}, \mathbf{Q}^i)} \left[ g((\mathbf{H}, \mathbf{Q}^i), u(\mathbf{H}, \mathbf{Q}^i)) \right.$$
$$\left. + \sum_{\mathbf{Q}^j} \Pr[\mathbf{Q}^j | (\mathbf{H}, \mathbf{Q}^i), u(\mathbf{H}, \mathbf{Q}^i)] \widetilde{V}(\mathbf{Q}^j) \right]$$
$$\stackrel{(c)}{\Rightarrow} \theta + \widetilde{V}(\mathbf{Q}^i) \qquad \forall 1 \leq i \leq I$$
$$= \mathbb{E} \left[ \min_{u(\mathbf{H}, \mathbf{Q}^i)} \left[ g((\mathbf{H}, \mathbf{Q}^i), u(\mathbf{H}, \mathbf{Q}^i)) \right.\right.$$
$$\left.\left. + \sum_{\mathbf{Q}^j} \Pr[\mathbf{Q}^j | (\mathbf{H}, \mathbf{Q}^i), u(\mathbf{H}, \mathbf{Q}^i)] \left( \sum_{\mathbf{H}'} \Pr[\mathbf{H}'] V(\mathbf{H}', \mathbf{Q}^j) \right) \right] | \mathbf{Q}^i \right]$$
$$\stackrel{(d)}{=} \min_{\mathbf{u}(\mathbf{Q}^i)} \widetilde{g}(\mathbf{Q}^i, \mathbf{u}(\mathbf{Q}^i)) + \sum_{\mathbf{Q}^j} \widetilde{f}(\mathbf{Q}^j | \mathbf{Q}^i, \mathbf{u}(\mathbf{Q}^i)) \widetilde{V}(\mathbf{Q}^j) \quad (26)$$

where (a) is due to (9), (b) is due to the definition $\widetilde{V}(\mathbf{Q}) \triangleq \mathbb{E}[V(\chi)|\mathbf{Q}]$, (c) is obtained by taking the conditional expectation (conditioned on $\mathbf{Q}^i$) on both sides of (10) and (d) is due to the definition of "conditional actions"

---

[12]The mild conditions refer to the two conditions in Theorem 1. One is the stepsize condition. The other is the condition on accessibility of the Markov Chain, which can be easily satisfied in most of the cases we are interested.

and $\widetilde{g}(\mathbf{Q}, \mathbf{u}(\mathbf{Q})) = \mathbb{E}[g(\boldsymbol{\chi}, u(\boldsymbol{\chi}))|\mathbf{Q}]$, $\widetilde{f}(\mathbf{Q}^j|\mathbf{Q}^i, \mathbf{u}(\mathbf{Q}^i)) = \mathbb{E}\Big[\Pr[\mathbf{Q}^j|\boldsymbol{\chi}^i, u(\boldsymbol{\chi}^i)]|\mathbf{Q}^i\Big]$.

## APPENDIX B: PROOF OF THEOREM 1

We shall first show the convergence of the martingale noise in (14). Let

$$q_k(\mathbf{Q}^i) \triangleq \mathbf{E}_k[Y_k(\mathbf{Q}^i)]$$
$$= T_i(\widehat{\mathbf{V}}_k(\mathbf{Q}^i)) - \widehat{\mathbf{V}}_k(\mathbf{Q}^i) - \big(T_I(\widehat{\mathbf{V}}_k(\mathbf{Q}^I)) - \widehat{\mathbf{V}}_k(\mathbf{Q}^I)\big) \quad (27)$$

$Y_k(\mathbf{Q}^i)$ is the noise-corrupt observation of $q_k(\mathbf{Q}^i)$. The martingale difference noise is

$$\delta M_k(\mathbf{Q}^i) = Y_k(\mathbf{Q}^i) - q_k(\mathbf{Q}^i)$$
$$= \big(\frac{S_{\widehat{g}_k}(i)}{l_k(i)} - \widetilde{g}(\mathbf{Q}^i)\big) - \big(\frac{S_{\widehat{g}_k}(I)}{l_k(I)} - \widetilde{g}(\mathbf{Q}^I)\big)$$
$$+ \big(\frac{S_{\widehat{V}_{k+1}}(i)}{l_{k+1}(i)} - \sum_{\mathbf{Q}^j} \widetilde{f}(\mathbf{Q}^j|\mathbf{Q}^i, \mathbf{u}(\mathbf{Q}^i))\widehat{V}_k(\mathbf{Q}^j)\big)$$
$$- \big(\frac{S_{\widehat{V}_{k+1}}(I)}{l_{k+1}(I)} - \sum_{\mathbf{Q}^j} \widetilde{f}(\mathbf{Q}^j|\mathbf{Q}^I, \mathbf{u}(\mathbf{Q}^I))\widehat{V}_k(\mathbf{Q}^j)\big)$$

with property that $\mathbf{E}_k[\delta M_k(\mathbf{Q}^i)] = 0$ and $\mathbf{E}[\delta M_k(\mathbf{Q}^i)\delta M_{k'}(\mathbf{Q}^i)] = 0, \forall k \neq k'$. For some $j$, define $M_k(\mathbf{Q}^i) = \sum_{l=j}^k \epsilon_l \delta M_l(\mathbf{Q}^i)$. Thus, from (14), we have

$$\widehat{\mathbf{V}}_{k+1}(\mathbf{Q}^i) = \widehat{\mathbf{V}}_k(\mathbf{Q}^i) + \epsilon_k\big[q_k(\mathbf{Q}^i) + \delta M_k(\widehat{\mathbf{V}}_k(\mathbf{Q}^i))\big]$$
$$= \widehat{\mathbf{V}}_j(\mathbf{Q}^i) + \sum_{l=j}^k \epsilon_l q_l(\mathbf{Q}^i) + M_k(\mathbf{Q}^i) \quad (28)$$

Since $\mathbf{E}_k[M_k(\mathbf{Q}^i)] = M_{k-1}(\mathbf{Q}^i)$, $M_k(\mathbf{Q}^i)$ is a martingale sequence. By martingale inequality, we have $\Pr_j\big\{\sup_{j \leq l \leq k} |M_l| \geq \lambda\big\} \leq \frac{\mathbf{E}_j[|M_k(\mathbf{Q}^i)|^2]}{\lambda^2}$. By the boundness assumption and the property of the martingale difference noise as well as the condition on the stepsize sequence, we have $\mathbf{E}_j[|M_k(\mathbf{Q}^i)|^2] = \mathbf{E}_j[|\sum_{l=j}^k \epsilon_l \delta M_l(\mathbf{Q}^i)|^2] = \sum_{l=j}^k \mathbf{E}_j[\epsilon_l^2 \delta M_l^2(\mathbf{Q}^i)] \leq \overline{M}\sum_{l=j}^k \epsilon_l^2 \Rightarrow \lim_{j \to \infty} \Pr_j\big\{\sup_{j \leq l \leq k} |M_l(\mathbf{Q}^i)| \geq \lambda\big\} = 0$. Thus, as $j \to \infty$, (28) goes to $\widehat{\mathbf{V}}_{k+1}(\mathbf{Q}^i) = \widehat{\mathbf{V}}_j(\mathbf{Q}^i) + \sum_{l=j}^k \epsilon_l q_l(\mathbf{Q}^i)$ with probability 1, the vector form of which is given by

$$\widehat{\mathbf{V}}_{k+1} = \widehat{\mathbf{V}}_j + \sum_{l=j}^k \epsilon_l \mathbf{q}_l$$
$$= \widehat{\mathbf{V}}_j + \sum_{l=j}^k \Big[\mathbf{T}(\widehat{\mathbf{V}}_l) - \widehat{\mathbf{V}}_l - \big(T_I(\widehat{\mathbf{V}}_l) - \widehat{V}_l(\mathbf{Q}^I)\big)\mathbf{e}\Big] \quad (29)$$

Next, we shall show the convergence of (29) after the martingale noise are averaged out. In the following proof, we use $i$ instead of $\mathbf{Q}^i$ for simplicity. Let $\Omega_k(i)$ denote the optimal control action attaining the minimum in $T_i(\widehat{\mathbf{V}}_k)$. Let $\widetilde{\mathbf{g}}_{\Omega_k}$ and $P_{\Omega_k}$ denote the conditional per-stage reward vector and conditional average transition probability matrix under the optimal control policy $\Omega_k$. Denote $w_k = T_I(\widehat{\mathbf{V}}_k) - \widehat{V}_k(I)$. We have

$$\mathbf{q}_k = \widetilde{\mathbf{g}}_{\Omega_k} + P_{\Omega_k}\widehat{\mathbf{V}}_k - \widehat{\mathbf{V}}_k - w_k\mathbf{e}$$
$$\leq \widetilde{\mathbf{g}}_{\Omega_{k-1}} + P_{\Omega_{k-1}}\widehat{\mathbf{V}}_k - \widehat{\mathbf{V}}_k - w_k\mathbf{e}$$
$$\mathbf{q}_{k-1} = \widetilde{\mathbf{g}}_{\Omega_{k-1}} + P_{\Omega_{k-1}}\widehat{\mathbf{V}}_{k-1} - \widehat{\mathbf{V}}_{k-1} - w_{k-1}\mathbf{e}$$
$$\leq \widetilde{\mathbf{g}}_{\Omega_k} + P_{\Omega_k}\widehat{\mathbf{V}}_{k-1} - \widehat{\mathbf{V}}_{k-1} - w_{k-1}\mathbf{e}$$
$$\Rightarrow \mathbf{A}_{k-1}\mathbf{q}_{k-1} - (w_k - w_{k-1})\mathbf{e} \leq \mathbf{q}_k$$
$$\leq \mathbf{B}_{k-1}\mathbf{q}_{k-1} - (w_k - w_{k-1})\mathbf{e}, \forall k \geq 1$$
$$\stackrel{\text{by iterating}}{\Rightarrow} \mathbf{A}_{k-1}\cdots\mathbf{A}_{k-m}\mathbf{q}_{k-m} - (w_k - w_{k-m})\mathbf{e} \leq \mathbf{q}_k$$
$$\leq \mathbf{B}_{k-1}\cdots\mathbf{B}_{k-m}\mathbf{q}_{k-m} - (w_k - w_{k-m})\mathbf{e}$$

From (27), we have $\mathbf{q}_k(I) = T_I(\widehat{\mathbf{V}}_k(I)) - \widehat{\mathbf{V}}_k(I) - \big(T_I(\widehat{\mathbf{V}}_k(I)) - \widehat{\mathbf{V}}_k(I)\big) = 0$ for all $k$. By the assumption (17), we have

$$(1-\delta)\min_{i'}\mathbf{q}_{k-m}(i') - (w_k - w_{k-m}) \leq \mathbf{q}_k(i)$$
$$\leq (1-\delta)\max_{i'}\mathbf{q}_{k-m}(i') - (w_k - w_{k-m}) \forall i$$
$$\Rightarrow \begin{cases}\min_{i'}\mathbf{q}_k(i') \geq (1-\delta)\min_{i'}\mathbf{q}_{k-m}(i') - (w_k - w_{k-m}) \\ \max_{i'}\mathbf{q}_k(i') \leq (1-\delta)\max_{i'}\mathbf{q}_{k-m}(i') - (w_k - w_{k-m})\end{cases}$$
$$\Rightarrow \max_{i'}\mathbf{q}_k(i') - \min_{i'}\mathbf{q}_k(i')$$
$$\leq (1-\delta)\big(\max_{i'}\mathbf{q}_{k-m}(i') - \min_{i'}\mathbf{q}_{k-m}(i')\big)$$
$$\Rightarrow \max_{i'}\mathbf{q}_k(i') - \min_{i'}\mathbf{q}_k(i') \leq \phi_j \prod_{l=1}^{\lfloor\frac{k-j}{m}\rfloor}(1-\delta_{j+lm})$$

where $\phi_j > 0$. Since $\mathbf{q}_k(I) = 0$, we have $\max_{i'}\mathbf{q}_k(i') \geq 0$ and $\min_{i'}\mathbf{q}_k(i') \leq 0$. Thus, $\forall i$, we have $|\mathbf{q}_k(i)| \leq \max_{i'}\mathbf{q}_k(i') - \min_{i'}\mathbf{q}_k(i') \leq \phi_j\prod_{l=1}^{\lfloor\frac{k-j}{m}\rfloor}(1-\delta_{j+lm})$.

Therefore, as $k \to \infty$, $\mathbf{q}_k \to \mathbf{0}$, i.e. $\widehat{\mathbf{V}}_k$ satisfies Bellman equation (18). By the Proposition 1 in Chapter 7 of [13], $J_\beta = T_I(\widehat{\mathbf{V}}_k(I)) - \widehat{V}_k(I)$ is the optimal value and $\widehat{\mathbf{V}}_k$ is the potential vector, which is up to an constant vector. However, due to the property that $\mathbf{q}_k(I) = 0 \forall k \Rightarrow \widehat{\mathbf{V}}_k(I) = \widehat{\mathbf{V}}_0(I) \forall k$, we have the convergence of the potential vector $\widehat{\mathbf{V}}_\infty = \lim_{k\to\infty}\widehat{\mathbf{V}}_k$ and the optimal value $J_\beta^* = T_I(\widehat{\mathbf{V}}_\infty) - \widehat{V}_\infty(\mathbf{Q}^I)$ by the online value iteration via stochastic approximation algorithm. Accordingly, the optimal stationary policy is $\Omega^* = \arg\min \mathbf{T}(\widehat{\mathbf{V}}_\infty)$.

## APPENDIX C: PROOF OF THEOREM 2

Solution of Bellman equation in (11) can be obtained by offline relative value iteration [19]. Without loss of generality, we set $\mathbf{Q}^I$ as the reference state. Hence, we have normalizing equation $\widetilde{V}^l(\mathbf{Q}^I) = 0 \forall l$. Assume $\widetilde{V}^l(\mathbf{Q}) = \sum_{k=1}^K \widetilde{V}_k^l(Q_k) \forall l$.

At the $(l-1)$-th iteration, updating policy according to (15) is equal to finding policy $\Omega_p^l$ which minimize the objective function (20). Under any given *CSI-only subcarrier allocation policy*, the optimal power actions for the $l$-th iteration which minimize (20) is given by

$$\tilde{p}_{k,n}^l(\mathbf{H}, Q_k) = \tilde{s}_{k,n}^l(\mathbf{H})\left(\frac{\frac{\tau}{\overline{N}_k}\Delta\widetilde{V}_k^{l-1}(Q_k)}{\gamma} - \frac{1}{|H_{k,n}|^2}\right)^+$$

$$\Rightarrow \overline{\mu}_k^l(Q_k) = \mathbb{E}\left[\sum_{n=1}^{N_F} \tilde{s}_{k,n}^l(\mathbf{H})\log(1+\tilde{p}_{k,n}^l(\mathbf{H},Q_k)|H_{k,n}|^2)|\mathbf{Q}\right]/\overline{N}_k$$

where $\overline{\mu}_k^l(Q_k)$ is the mean departure rate under $\mathbf{u}_k^l(Q_k) = \{u_k^l(\mathbf{H},Q_k) : (\mathbf{H},Q_k)\forall\mathbf{H}\}$ with $u_k^l(\mathbf{H},Q_k) = (\tilde{p}^l(\mathbf{H},Q_k),\tilde{s}^l(\mathbf{H}))$, and $\Delta\widetilde{V}_k^l(Q_k) = \widetilde{V}_k^l(Q_k) - \widetilde{V}_k^l([Q_k-1]^+)$ is the potential increment for the $k$-th queue.

At the $l$-th iteration, we determine the potential $\widetilde{V}^l(\mathbf{Q})$ and $\theta^l$ by solving the normalizing equation $\widetilde{V}^l(\mathbf{Q}^I) = 0$ together with $I = (N_Q+1)^K$ Poisson equations

$$\theta^l + \widetilde{V}^l(\mathbf{Q}^i) = \sum_k \tilde{g}_k(Q_k^i, \mathbf{u}_k^l(Q_k^i)) + \widetilde{V}^l(\mathbf{Q}^i)$$
$$+ \sum_k \lambda_k \tau \Delta\widetilde{V}^l(Q_k^i+1) - \sum_k \overline{\mu}_k^l(Q_k^i)\tau\Delta\widetilde{V}^l(Q_k^i)$$

$$\Longrightarrow \theta^l = \sum_k \theta_k^l, \quad \forall 1 \le i \le I \tag{30}$$

where

$$\theta_k^l \qquad \forall 0 \le Q_k \le N_Q \tag{31}$$
$$= \tilde{g}_k(Q_k^i, \mathbf{u}_k^l(Q_k^i)) + \lambda_k\tau\Delta\widetilde{V}_k^l(Q_k^i+1) - \overline{\mu}_k^l(Q_k^i)\tau\Delta\widetilde{V}_k^l(Q_k^i)$$

$$\tilde{g}_k(Q_k^i, u_k^l(Q_k^i)) = \beta_k \frac{Q_k^i}{\lambda_k} + \gamma\mathbb{E}\left[\sum_n \tilde{p}_{k,n}^l(\mathbf{H}_n, Q_k^i)|Q_k^i\right]$$

There are $I$ joint $\mathbf{Q} = (Q_1,\cdots,Q_K)$ states, but there are only $N_Q+1$ states for $Q_k\ \forall k$. Hence, in the original $(N_Q+1)^K$ Poisson equations (30), there are only $N_Q+1$ independent Poisson equations in (31) for the $k$-th ($1\le k \le K$) queue. In addition, set $\widetilde{V}_k^l(0) = 0\ \forall k$ as the individual normalizing equation, which also satisfies $\widetilde{V}^l(\mathbf{Q}^I) = \sum_k \widetilde{V}_k^l(0) = 0$. Hence, in the $l$-th iteration, we can obtain $\{\widetilde{V}_k^l(Q_k),\theta_k^l\}$ by solving the $k$-th user's reduced state Poisson equation in (31) together with its the individual normalizing equation. Accordingly, $\{\widetilde{V}^l(\mathbf{Q}),\theta^l\}$ is the solution for the original $(N_Q+1)^K$ Poisson equations (30), where $\widetilde{V}^l(\mathbf{Q}) = \sum_k \widetilde{V}_k^l(Q_k)$ and $\theta^l = \sum_k \theta_k^l$.

Continue the iterations until $\Omega_p^{(l+1)} = \Omega_p^{(l)}$. We obtain $\{\widetilde{V}_k(Q_k), \theta_k\}$, which is the solution of the $k$-th user's reduced Bellman equation in (23). Accordingly, $\{\widetilde{V}(\mathbf{Q}),\theta\}$ is the solution of the original $(N_Q+1)^K$ Bellman equations in (11), where $\widetilde{V}(\mathbf{Q}) = \sum_k \widetilde{V}_k(Q_k)$ and $\theta = \sum_k \theta_k$, which are the potential for the joint $\mathbf{Q}$ and the optimal average reward respectively.

## APPENDIX D: PROOF OF COROLLARY 1

Since $\{\widetilde{V}_k(Q_k),\theta_k\}$ is the solution of the $k$-th user's reduced Bellman equation (23), the original MDP can be decoupled into $K$ individual MDP's with transition kernel is given by $\tilde{f}(Q_k+1|Q_k, \mathbf{u}_k(Q_k)) = \lambda_k\tau$, $\tilde{f}(Q_k-1|Q_k,\mathbf{u}_k(Q_k)) = \overline{\mu}_k(Q_k)\tau$ and $\tilde{f}(Q_k|Q_k,\mathbf{u}_k(Q_k)) = 1 - \lambda_k\tau - \overline{\mu}_k(Q_k)\tau$. Therefore, the online value iteration can be applied to $K$ individual MDP's respectively. Under the same condition of Theorem 1[13], almost sure convergence of the online value iteration algorithm in Corollary 1 is guaranteed.

Next, we have to show that the converged solution is indeed asymptotically optimal. Denote $k_n^* \triangleq \arg\max_k |H_{k,n}|^2$. For large $K$, $|H_{k_n^*,n}|^2$ grows with $\log(K)$ by extreme value theory. Because the traffic loading remains unchanged as we scale up $K$, $\max_{k,j}|\Delta_k\widetilde{V}(\mathbf{Q}) - \Delta_j\widetilde{V}(\mathbf{Q})| = O(1)$. Hence, $X_{k_n^*,n}$ grows like $\log(\log(K))$. As $K \to \infty$, $\Pr[k_n^* = \arg\max_k x_{k,n}] = 1$. Thus the subband allocation result of optimal subband allocation in (22) and the best CSI subband allocation in (24) will be the same for large $K$. Thus, (24) and (25) are asymptotically optimal. Therefore, with (24), following the proof of Theorem 2, we can prove $\sum_k \widetilde{V}_k(Q_k) \to \widetilde{V}^*(\mathbf{Q})$, $\sum_k \theta_k \to \theta^*$ as $K \to \infty$, where $\widetilde{V}^*(\mathbf{Q})$ and $\theta^*$ are the potential and optimal average reward under the global optimal power and subcarrier allocation given by Lemma 2.

---

[13]It can be easily verify that all states in the birth-death chain are accessible.

IEEE TRANSACTIONS ON WIRELESS COMMUNICATIONS, VOL. 9, NO. 1, JANUARY 201011

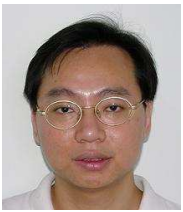
**Vincent K. N. Lau** obtained B.Eng (Distinction 1st Hons) from the University of Hong Kong in 1992 and Ph.D. from Cambridge University in 1997. He was with PCCW as system engineer from 1992-1995 and Bell Labs - Lucent Technologies as member of technical staff from 1997-2003. He then joined the Department of ECE, HKUST as Associate Professor. His current research interests include the robust and delay-sensitive cross-layer scheduling, cooperative and cognitive communications as well as stochastic approximation and Markov Decision Process.

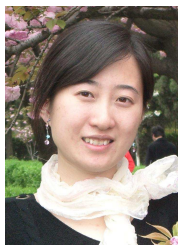
**Ying Cui** received B.Eng degree (first class honor) in Electronic and Information Engineering, Xian Jiaotong University, Xi'an, China in 2007. She is currently a Ph.D candidate in the Department of Electronic and Computer Engineering, the Hong Kong University of Science and Technology (HKUST). Her current research interests include cooperative and cognitive communications, delay-sensitive cross-layer scheduling as well as stochastic approximation and Markov Decision Process.